 \documentclass[sigconf,screen=true]{acmart}

\AtBeginDocument{%
  \providecommand\BibTeX{{%
    \normalfont B\kern-0.5em{\scshape i\kern-0.25em b}\kern-0.8em\TeX}}}

\copyrightyear{2020}
\acmYear{2020}
\setcopyright{acmlicensed}\acmConference[IVA '20]{IVA '20: Proceedings of the 20th ACM International Conference on Intelligent Virtual Agents}{October 19--23, 2020}{Virtual Event, Scotland Uk}
\acmBooktitle{IVA '20: Proceedings of the 20th ACM International Conference on Intelligent Virtual Agents (IVA '20), October 19--23, 2020, Virtual Event, Scotland Uk}
\acmPrice{15.00}
\acmDOI{10.1145/3383652.3423911}
\acmISBN{978-1-4503-7586-3/20/09}


\acmConference[IVA '20]{IVA '20: 20th International Conference on Intelligent Virtual Agents}{October 19--23, 2020}{Glasgow, United Kingdom}

\usepackage{hyperref}

\usepackage{acronym}

\usepackage[linesnumbered,ruled,vlined]{algorithm2e}

\newacro{VAD}{Voice Activity Detection}

\begin{document}


\title[Let's Face It]{Let's Face It: Probabilistic Multi-modal Interlocutor-aware Generation of Facial Gestures in Dyadic Settings}


\author{Patrik Jonell}
\affiliation{KTH Royal Institute of Technology}
\email{pjjonell@kth.se}

\author{Taras Kucherenko}
\affiliation{KTH Royal Institute of Technology}
\email{tarask@kth.se}

\author{Gustav Eje Henter}
\affiliation{KTH Royal Institute of Technology}
\email{ghe@kth.se}

\author{Jonas Beskow}
\affiliation{KTH Royal Institute of Technology}
\email{beskow@kth.se}


\begin{abstract}
%  for example the facial gesturing (expressions and head movements)

To enable more natural face-to-face interactions, conversational agents need to adapt their behavior to their interlocutors. One key aspect of this is generation of appropriate non-verbal behavior for the agent, for example facial gestures, here defined as facial expressions and head movements. Most existing gesture-generating systems do not utilize multi-modal cues from the interlocutor when synthesizing non-verbal behavior. Those that do, typically use deterministic methods that risk producing repetitive and non-vivid motions. In this paper, we introduce a probabilistic method to synthesize interlocutor-aware facial gestures -- represented by highly expressive FLAME parameters -- in dyadic conversations. Our contributions are: a) a method for feature extraction from multi-party video and speech recordings, resulting in a representation that allows for independent control and manipulation of expression and speech articulation in a 3D avatar; b) an extension to MoGlow, a recent motion-synthesis method based on normalizing flows, to also take multi-modal signals from the interlocutor as input and subsequently output interlocutor-aware facial gestures; and c) a subjective evaluation assessing the use and relative importance of the different modalities in the synthesized output.
The results show that the model successfully leverages the input from the interlocutor to generate more appropriate behavior. Videos, data, and code are available at: \href{https://jonepatr.github.io/lets\_face\_it}{https://jonepatr.github.io/lets\_face\_it}.

\end{abstract}

\maketitle

\begin{figure}[t]
\begin{center}
\includegraphics[width=0.3\textwidth]{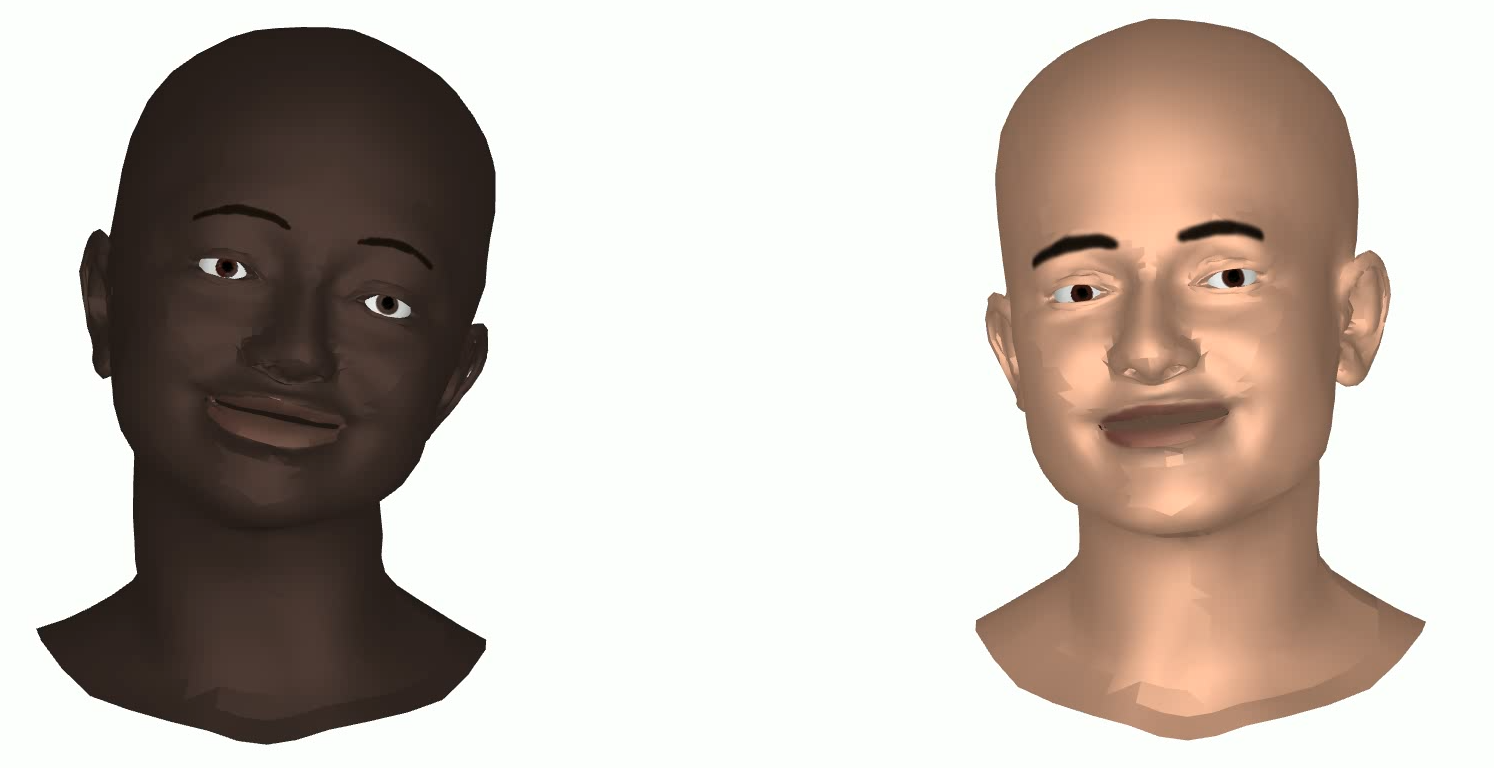}
\vspace{-1.5ex}
\caption{Two avatars in a snapshot from our experiments.}
% \vspace{-4mm}
\label{fig:intro}
\end{center}
\vspace{-9mm}
\end{figure}

\section{Introduction}
Generating appropriate facial gestures (here defined as facial expressions and head movements) for a conversational agent in a dyadic setting is a task as intriguing as it is challenging. Its usefulness in human-agent interaction has been researched extensively \cite{luo2013examination,doi:10.1111/j.1467-9280.2005.01619.x,nass1994computers} and there have been many attempts at realizing its potential in both virtual agents \cite{kucherenko2019analyzing, sadoughi2018novel} and social robots \cite{yoon2018robots}.
It is well known that facial motion is highly correlated with speech, and often contains cues that contribute to or reinforce the spoken message \cite{ekman2004emotional}. But facial expressions in a dyadic setting are also strongly affected by the other party. Interpersonal dynamics in face-to-face conversation includes many phenomena that affect the interaction in different ways, such as mimicry -- the tendency to adopt poses, facial expressions, mannerisms, and speaking styles of the interlocutor. 
%Mimicry has been referred to as a social glue \cite{lakin2003chameleon} and has been shown to increase learning and attention in tutoring situations \cite{martin2012mimicry}.
For example, it has been shown that when a conversational agent just copies the facial expressions of the human interlocutor with some delay it is perceived as more trustworthy \cite{doi:10.1111/j.1467-9280.2005.01619.x}. Furthermore, Cassell and Thorisson \cite{cassell1999power} found that so-called envelope feedback (e.g. gaze, manual beat gestures, and head movements) to be more important for the user than emotional feedback when interacting with conversational agents. 
% Although there are many systems which use multi-modal input from the interlocutor, most of these systems often adapt and change only the semantic output of the system rather than the non-verbal output; some examples are \cite{leite2011modelling, ritschel2017adapting}.
As modeling conversational dynamics is difficult to achieve, most non-verbal behavior generation methods only use speech and/or semantic content produced by the agent as inputs to the system \cite{jonell2019learning,kucherenko2019analyzing,yoon2018robots}.
%MoGlow~\cite{alexanderson2020style} is an example of a system which produces (full body) gestures in a probabilistic fashion, leading to non-repetitive, natural looking motion, but unaware of any interlocutor. 
Recently a few systems have been introduced that use non-verbal behaviors from the interlocutor to control non-verbal output from the system \cite{huang2017dyadgan,ahuja2019react,feng2017learn2smile,greenwood2017dyadic}. We continue this line of work and present a probabilistic system, based on normalizing flows, for generating facial gestures in dyadic settings. Our system takes in audio from both conversational partners and facial gestures of the interlocutor and generates corresponding appropriate facial gestures for the virtual agent in a given context.

We evaluate this system using segments annotated as containing mimicry from a database of dyadic interactions, these being salient examples of interlocutor-dependent non-verbal behavior. Our stimulus-generation method allowed manipulating speech articulation independently from the facial gestures, allowing for varying the facial gestures while controlling for the effect of speech context.
%These segments have been created such that the speech articulation is independently controlled from the facial gestures, allowing for controlling the speech context while varying the facial gesturing.
We find that:
% terms of mimicry through a series of four studies, where we show that 
(1) Evaluators can distinguish mimicry segments from mismatched segments (from the same interaction but another point in time) and find mimicry segments more appropriate. This also validates that our feature extraction and stimulus generation methods are appropriate for non-verbal behavior.
%ly captures non-verbal behavior.
(2) Feeding our model mismatched input segments yields a less appropriate response to the interlocutor, showing that our model leverages the multi-modal signals from the interlocutor to generate more appropriate facial gestures. 
(3) Removing the interlocutor's facial gestures as input led to less appropriate behavior, while interlocutor speech was not beneficial for facial-gesture generation in our scenario.

In order for researchers to build on top of our work, our extracted database of features and analysis-synthesis code can be found on the project website: \href{https://jonepatr.github.io/lets\_face\_it}{https://jonepatr.github.io/lets\_face\_it}.
\vspace{-1.5ex}

\section{Related work}

\subsection{Representing facial communicative signals}

While there are many methods for representing facial communicative signals, our scenario and experiments impose the following requirements:
Firstly, we require a parameterization that allows encoding facial gestures from video (to be used as inputs and output to the models) in a person-independent way.
Secondly, we need to generate an animated 3D avatar, so we require a reliable inversion of the parameterization to render faces that express the perceptually relevant elements.
Finally, we need independent control over speech articulation and facial expression, in order to be able to run experiments with out-of-context gestures as in Section \ref{sec:evaluation}.

Ekman \& Friesen's Facial Action Coding System (FACS) \cite{ekman1978facial} was developed for subject-independent coding of facial expressions for psychology research. It has also been widely used in graphics and machine-learning applications \cite{chu2018face,10.1007/978-3-030-21902-4_5,dermouche2019generative}, but while FACS is well suited for coding, e.g., emotional expressions, it is less ideal for speech animation. There is also no canonical way of automatically encoding and decoding between video and FACS.

Another commonly used parametrization is facial landmarks, for example the 68 point Multi-PIE scheme, e.g., used in \cite{feng2017learn2smile}. Facial landmarks often lack resolution and are not fully able to represent facial expressions and emotions \cite{liu2013representing}. They also lead to subject-specific data and cannot easily be used in generation. MPEG-4 Face Animation Parameters (FAP) are closely related to FACS but were designed to cope with both analysis and synthesis and are, for example, used as output parameters in
%Dermouche et al.'s work
\cite{dermouche2019generative}. There is however a lack of reliable tools for reconstruction/synthesis. Statistically-based 3D analysis/synthesis parameterizations such as 3D morphable models \cite{blanz2003face} and Active Appearance Models \cite{cootes2001active} can yield high-quality results, but they typically rely on manual initialization steps that make them expensive to deploy in large-scale multi-talker machine learning settings with many hours worth of data. 

FLAME \cite{FLAME:SiggraphAsia2017} is a new parameterization that
%is able to
represents facial expressions, shapes, and head rotation in a low-dimensional Principal Component Analysis (PCA) parameter space
%that can be realized
realizable as a 3D mesh. Expression parameters can be automatically extracted from video. Our system uses FLAME parameters as this improves the fidelity of facial gestures. FLAME allows independent control over expression and shape by design. Using techniques described in Section \ref{sec:stimuli_generation} it is furthermore possible to independently drive speech articulation and facial expression.

\subsection{Gesture generation}

% Researchers have investigated using the \textit{emotional state} of the user as an input parameter to adaptive social robot \cite{ahmad2017adaptive,8489158}. \textit{Emotional state} has also been used in commercial robots, such as the robot Pepper which is able to recognize the users \textit{emotional state} from the voice and facial expressions and adapt its behavior accordingly \cite{pepper}.

% There are numerous physiological signals that researchers have been using for affective state modeling, for example, Picard et al. used facial muscle  tension, blood volume pressure, skin  conductance and respiratory rate \cite{picard2001toward}. Boccanfuso et al. used skin temperature for classifying the user's emotional state \cite{boccanfuso2016thermal}. 

% \subsection{Facial gestures in conversational agents}
% Researchers have for a long time tried to both understand but also realize adequate facial gesturing behavior between a conversational agent and a human interlocutor. Cafaro et al. explored first impressions of a virtual agent by manipulating smile, gaze and proximity \cite{10.1007/978-3-642-33197-8_7}, and the results indicated that it had an effect on the perception of the virtual agent. 
% There have been attempts at realizing such systems, by for example creating rule-based systems [\cite{douville1998rule}]. 

% 10.1007/978-3-319-21996-7_16

Several previous works have demonstrated successful generation of gestures of various kinds.
Recent work in speech-driven hand-gesture generation, for example, has primarily been based on deep learning.
Hasegawa et al.~\cite{hasegawa2018evaluation} designed a neural network to map from speech audio to 3D motion sequences. Kucherenko et al.~\cite{kucherenko2019analyzing} extended this work to learn a better representation of the motion, achieving smoother gestures as a result. Yoon et al.~\shortcite{yoon2018robots} learned a mapping from text to gestures using a recurrent neural network.
%Habibie et al.~\cite{habibie2017recurrent} applied Variational Autoencoders (VAEs)~\cite{kingma2013auto} for motion generation. They encoded a sequence of control signals into a latent representation using a CNN-based encoder, then used a Long Short-Term Memory (LSTM) decoder to synthesize motion. 
% , similarly
Speech-driven head-motion and facial gesture generation has been performed using methods such as Variational Autoencoders (VAEs) \cite{kingma2013auto} to predict head pose conditioned on acoustic features \cite{greenwood2017predicting}, Bidirectional Long Short-Term Memory (BLSTM) networks \cite{greenwood2017dyadic,greenwood2018joint,sadoughi2017joint}, and conditional Generative Adversarial Networks (GANs) \cite{goodfellow2014generative} as seen in \cite{sadoughi2018novel,chu2018face}.
% incorporating BLSTMs \cite{sadoughi2018novel}. 
% Similarly, facial expression generation has been explored by using BLSTMs \cite{sadoughi2017joint} and GANs \cite{chu2018face}.
In another line of work, Karras et al.~\cite{karras2017nvidia} trained a CNN-based neural network using speech together with a learned emotion representation as input to generate corresponding 3D meshes of faces with impressively little training data. 

\subsection{Interlocutor-aware gesture generation}

Our problem formulation is largely inspired by a recent method to model conversational dynamics for gesture generation \cite{ahuja2019react}. Like in that work, we also model avatar behavior based on both the avatar's own speech and the speech and motion of the interlocutor. One main difference between our work and that paper is that we model a different aspect of non-verbal behavior, namely facial gestures instead of hand gestures. Another important difference is that our method is not deterministic, but probabilistic. Their method is also based on data from motion capture, while our system uses regular videos as input and extracts features from these videos.

One similar work that uses a probabilistic method is DyadGAN \cite{huang2017dyadgan}, which trained a conditional GAN to generate face images based on the interlocutor's facial expressions. However, the work only produced a single image, ignoring temporal aspects. DyadGAN was later extended to generate sequences of interlocutor-aware facial gestures \cite{nojavanasghari2018interactive}. However, they did not use speech information, nor did they produce output parameters that can control a virtual agent.

Feng et al.~\cite{feng2017learn2smile} presented a system using VAEs to generate facial gestures. 
%Their system, however, is uni-modal and only uses facial information from the interlocutor, while our proposed method is multi modal. 
However, their system is limited to sequences of facial gestures already existing in the training dataset, while our system is able to generate completely new motions. Furthermore, our system also relies on FLAME parameters for parametrization of the facial features as opposed to facial landmarks, granting several benefits; most importantly, the output parameters can directly generate a high-quality 3D face with corresponding gestures while simultaneously providing independent control over lip-sync and facial shape. Dermouche et al.~\cite{dermouche2019generative} presented a system similar to Feng et al.\ but added the conversational state as additional conditional information, and also created a system usable in real time. They encoded the input using LSTMs while outputting FAPs. 
% After conducting an experiment with the real-time system at a museum, their conclusion was that the system was preferred by the users if it smiled when the users smiled \cite{dermouche2019generative}.

\subsection{Normalizing flows}
In this work we use normalizing flows \cite{papamakarios2019normalizing} for probabilistic modeling. This has several advantages over other methods such as VAE or GANs, as detailed in \cite{henter2019moglow}.
% Essentially, normalizing flows offer the best of both worlds, combining the power and flexibility of GANs with easy training based on exact maximum likelihood, like in classic probabilistic models such as mixture densities \cite{bishop1994mixture}.
The specific model we use is adopted from MoGlow \cite{henter2019moglow}, which adapted a normalizing-flow method called Glow \cite{kingma2018glow} to the problem of motion generation.
%based on , which is a probabilistic generative model from the Normalizing Flow family.
We describe the MoGlow method more in detail in Section \ref{sec:sys}. The method has been successfully applied to gesture generation \cite{alexanderson2020style}, which inspired us to apply it to our problem as well. However, our system differs from MoGlow as we use several modalities to condition the model, each encoded by a separate neural network, and we apply the model to another task (interlocutor-aware facial-gesture generation). Another difference from both MoGlow and Ajuha et al.'s work \cite{ahuja2019react} is that we start from regular monocular videos, thus not requiring data recorded using specialized motion-capture equipment.

\section{System architecture}
\label{sec:sys}

\subsection{Problem formulation}

We frame the problem of generating interlocutor-aware facial gestures in the following way:
%similar to the recent method introduced by Ahuja et al. \cite{ahuja2019react}: 
given a sequence of speech features of the avatar $\boldsymbol{S^a} = [\boldsymbol{s^a}_t]_{t=1:T}$ as well as the interlocutor's facial gestures $\boldsymbol{F^i} = [\boldsymbol{f^i}_t]_{t=1:T}$  and speech features $\boldsymbol{S^i} = [\boldsymbol{s^i}_t]_{t=1:T}$,
the task is to generate a corresponding facial gesture sequence $\boldsymbol{\hat{F^a}} = [\boldsymbol{\hat{f^a}}_t]_{t=1:T}$ that the avatar might perform in the conversation.

\subsection{Model foundations}
%In this section we briefly describe the key methods we use in our work.

The model we utilize to generate motion in this work belongs to the class of probabilistic generative models called \emph{normalizing flows}.
Normalizing flows are similar to GANs in that they generate output by drawing samples from a simple \emph{base} or \emph{latent distribution} $\boldsymbol{Z}$ (here a standard normal distribution) and then transform these samples nonlinearly using a neural network $\boldsymbol{g}$ such that the transformed output distribution $\boldsymbol{X}=\boldsymbol{g}(\boldsymbol{Z})$ matches that of the data.
Different from the one-way neural networks in GANs, however, normalizing flows use \emph{invertible} nonlinear transformations, so called \emph{invertible neural networks}, for $\boldsymbol{g}$.
The approach gains power and expressivity by chaining together several simple nonlinear transformations, called \emph{steps} of flow, analogous to the layers in a regular neural network.
%The invertibility makes it possible to use the change-of-variables formula to compute the exact likelihood of training data examples.
% Normalizing flow models (in practice, the weights of the invertible network $\boldsymbol{g}$) can therefore be trained through gradient-based optimization minimizing the negative log-likelihood (NLL) of minibatch data.
%, similar to likelihood maximization with classic probabilistic models such as hidden Markov models (HMMs) \cite{rabiner1989tutorial} and mixture density networks (MDNs) \cite{bishop1994mixture}. %,fragkiadaki2015recurrent}.
% This means that flows can straightforwardly train generative models with an expressive power similar to GANs but without using a discriminator, thus avoiding the theoretical \cite{mescheder2018training} and practical \cite{lucic2018gans} issues that complicate GAN training.
For more details on normalizing flows please see \cite{papamakarios2019normalizing}.

The model in this paper is based on a specific normalizing flow transformation $\boldsymbol{g}$ called Glow \cite{kingma2018glow}.
This choice allows both fast likelihood computation and efficient sampling from the learned distribution.
Our model structure is similar to the MoGlow architecture used for autoregressive generation of pose sequences in 
%a previous model called MoGlow \cite{henter2019moglow}, that adapts Glow to generate a sequence of poses in an autoregressive manner and has demonstrated impressive results in both
locomotion \cite{henter2019moglow} and gesture generation \cite{alexanderson2020style}.
These papers also show how the nonlinear transformation $\boldsymbol{g}$, and thus the learned distribution $\boldsymbol{X} = \boldsymbol{g}(\boldsymbol{Z})$, can be made to depend on conditioning information that affects the motion, including an external control signal.
%In practice, invertible neural networks $\boldsymbol{g}$ use several regular (one-way) neural networks as building blocks internally.
%Like in MoGlow, we use recurrent neural networks in the time direction for these neural networks (for long memory and stability). We also feed in the conditioning information we have extracted from (...) as additional inputs to these networks, in order to make the transformation, and thus the learned distribution $\boldsymbol{X} = \boldsymbol{g}(\boldsymbol{Z})$" depend on the interlocutor and the agent's own speech.
Specifically, MoGlow feeds the conditioning information as an additional input to the regular (one-way) neural networks contained inside each step of flow (see \cite{henter2019moglow}).
We will use this control signal to create models of non-verbal behavior that are able to use the interlocutor's speech and facial gestures.
Like in MoGlow, we do not use any hierarchical structure in the generator, meaning that $L=1$ in the language of Kingma et al.~\cite{kingma2018glow}.

\subsection{Proposed model overview}

\begin{figure}%[ht]
\begin{center}
\includegraphics[width=0.9\columnwidth]{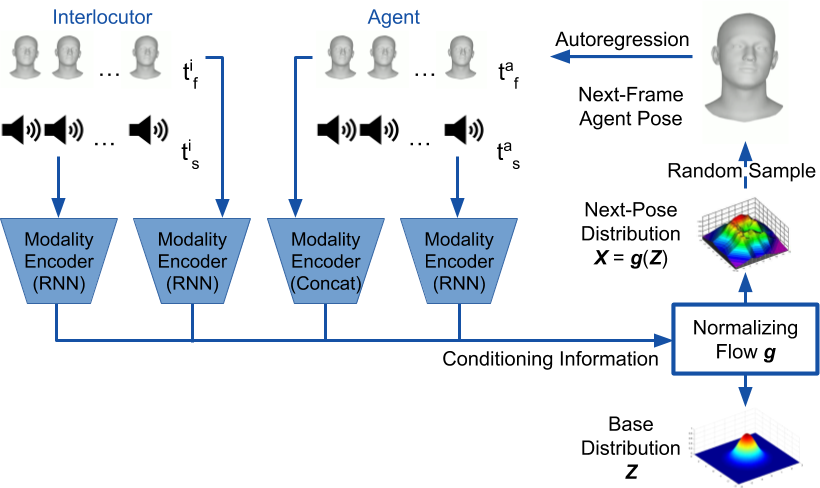}
% \vspace{-1ex}
\caption{System architecture. While we visualize conversation parties as talking heads in the figure, the facial gesture inputs and outputs of the machine-learning system were FLAME parameters. Similarly, audio inputs were MFCCs and prosodic features, rather than raw waveforms.}
\vspace{-2ex}
\label{fig:system}
\end{center}
% \vspace{-3mm}
\end{figure}

Our model generates facial gestures conditioned on the speech of the avatar as well as the speech and the facial features of the interlocutor. A graphical overview of the model is shown in Figure~\ref{fig:system}. The core of the model is the normalizing flow, which transforms Gaussian driving noise (shown below the model) into a distribution of facial expressions (shown on top of the model). In order to be able to generate smooth facial motion, the model is made autoregressive -- it uses the avatar's facial expressions from preceding frames as an extra conditioning to generate the next frame. The generated facial motion should be consistent with the avatar's speech (but not necessarily its semantics) and hence our model is also conditioned on the avatar's speech signal from previous $t^a_s$ time-steps. To enable generating appropriate behavior toward the interlocutor, the speech and facial motion of the interlocutor for the $t^i_s$ and $t^i_f$ time-steps, respectively, are used as additional conditioning for the normalizing flow. The proposed model hence learns to generate a distribution of appropriate facial gestures using multi-modal conditioning.

Since no previous facial expressions are available at test time, the model starts generation with a sequence of zero vectors standing in for the missing facial-gesture inputs.
%(determine by the history hyperparameter).

Like in MoGlow \cite{henter2019moglow} the conditioning information is concatenated with the other inputs to the networks inside the steps of flow, but in our system each modality is encoded by a separate network (and later subjected to an additional transformation which is different for each step),
% and joined together by a linear layer  before concatenating to the affine coupling layer
as described in the next subsection.

\subsection{Modality encoder}
Four different inputs are used in our model to condition the output distribution: the interlocutor's acoustic and facial features, as well as the agent's own acoustic features and previous facial features (as autoregressive input to ensure continuity). How the acoustic and facial features were extracted is described in Section~\ref{sec:feature_extraction}. Below we describe our modality encoders shown in Figure~\ref{fig:system}.

We experimented with different neural networks for encoding each modality: Multi-layer Perceptrons (MLPs), Recurrent Neural Networks (RNNs) and 1D-convolution networks (CNNs). We decided on the final configuration (RNNs) based on an initial hyperparameter search on the validation dataset.
However, for the autoregression, the avatar's previous facial features were passed into the normalizing flow model without any processing: simply as a concatenation of $t^a_{p_f}$ previous frames. All other modalities were first encoded from input histories of a given time duration (different for different modalities) into fixed-length vectors using separate RNNs, specifically using Gated Recurrent Units (GRUs)~\cite{cho2014learning}. We took both the hidden state and the final output from the GRUs to retain more information.
%All conditioning information was used by every step of the normalizing flow in the following way:
For each step of flow, all modality encodings were concatenated and then passed through a one-layer neural network with a LeakyReLU activation function. This transformation network was different in every step of the flow, resulting in different conditioning vectors in each step. The per-step conditioning information was used to influence the transformation in each step in the same way as in MoGlow.

\subsection{Training scheme}
\label{sec:training_scheme}
We used teacher forcing without annealing or scheduled sampling. This means that the model always received the ground-truth autoregressive input during training instead of samples from the model, since the latter can make models converge on incorrect output \cite{huszar2015not}.

We used the Adam optimizer \cite{kingma2014adam} since it has been used before to train similar systems \cite{henter2019moglow,alexanderson2020style}. We also used learning-rate warm up, as is common for normalizing flows \cite{kingma2018glow}.
Different learning-rate schedulers were tested, but did not seem to impact the results.

% \begin{algorithm}%[H]
%  \KwData{Training dataset}
%  \KwResult{Trained model}
%  initialize the model and $nll_{d}=inf$\;
%  \For{n epochs}{
%      \For{each training batch}
%     {
%       sample p from uniform [0,1]\;
%       \eIf{$p<0.1$ and $nll_{d}>0$}{
%           generate $permutation\_indices$ \;
%           permute $F_i$ and $S_i$ with these permutation:\\
%           $\hat{F}_i = permute(F_i,  permutation\_indices)$ ;\\ 
%           $\hat{S}_i = permute(S_i,  permutation\_indices)$;\
%           calculate negative log - likelihood:\; 
%           $nll_{d} = nll(model(F_a|S_a, \hat{F}_i, \hat{S}_i))$\;
%           $loss = - nll_{d} * 0.2 $ \;
%       }{
%           $loss = nll(model(F_a|S_a, F_i,S_i))$\;
%       }
%       gradient = Adam(model, loss)\;
%       model.update(model, gradient)\;
%     }
%     \For{each validation batch}
%     {
%     repeat steps 6--10
%     }
%  }
%  \caption{Our training scheme. Negative log-likelihood is denoted $nll$. Mismatched NLL is denoted $nll_d$}
%  \label{alg:learning}
% \end{algorithm}

% \begin{figure*} %[ht]
% \begin{center}
% \includegraphics[width=0.69\textwidth]{pics/feature_extraction_v6.png}
% \caption{Feature extraction. Do we need this?}
% \label{fig:data}
% \end{center}
% \end{figure*}

In order for the model to listen more to the conditioning from the interlocutor we used a special training scheme based on negative learning \cite{munawar2017limiting}.
% and shown in Algorithm~\ref{alg:learning}. 
The main idea is to not only minimize the loss of the training examples, but also maximize the loss of ``wrong'', negative examples. There was a 0.1 probability to use a negative sample for each batch. Negative samples are created by shuffling both facial $\boldsymbol{F^i}$ and speech conditioning $\boldsymbol{S^i}$ in the conditioning information for the whole batch, so that each output sequence in the batch now has the conditioning information of a different sample. Temporal consistency was preserved -- the mismatched conditioning was still a continuous sequence but from another example. Mathematically, a permutation of elements where no element appears in its original position is known as a \emph{derangement}, but we will refer to such samples with deliberate incorrect conditioning as \emph{mismatched}.

In order to make the model better at distinguishing between appropriate from inappropriate output motion, we want the log-likelihood for mismatched samples to be as small as possible. We therefore switch the sign of the log-likelihood of negative examples. This was done as long as the negative log-likelihood (which we use as the loss in these cases) was positive for those negative examples, an occurrence that became increasingly rare as the loss kept decreasing as the model improved during training.
%, which was also evaluated during the validation phase.

\subsection{Implementation and hyperparameters}
% \footnote{\href{https://pytorch.org}{pytorch.org}}
Our implementation used the PyTorch-based GitHub repository glow-pytorch\footnote{\href{http://github.com/chaiyujin/glow-pytorch}{https://github.com/chaiyujin/glow-pytorch}} %chaiyujin/glow-pytorch
 as a base, adapted to PyTorch Lightning\footnote{\href{https://github.com/PyTorchLightning/pytorch-lightning}{https://github.com/PyTorchLightning/pytorch-lightning}}
 %~\cite{falcon2019pytorch}.
The hyperparameter search used Optuna \cite{optuna_2019}, which identified the following hyperparameters that we used in our experiments for the proposed model: total conditioning dimensionality = 512, initial learning rate = $10^{-5}$, training sequence length = 80.
The Glow parameters were K~=~16~steps of flow with 128 hidden channels. All other hyperparameters of the final model can be found on the project website. The final model was trained for 15 epochs on a single GPU for approximately 40 hours.

% \begin{table}[]
% \caption{Encoder hyperparameters. History indicates how many previous frames of a given modality that were used.}
% \vspace{-1ex}
% \label{tab:enc_hparams}
% \begin{tabular}{@{}lllll@{}}

% \hline
% Modality            & Enc.\ Dim. & Enc.\ Type & History Len. & Dropout \\
% $\boldsymbol{S^a}$  & 128           & RNN           & 2              & 0.5     \\
% $\boldsymbol{F^a}$  & 256           & Concat        & 5              & 0       \\
% $\boldsymbol{S^i}$  & 256           & RNN           & 16             & 0.3  \\
% $\boldsymbol{F^i}$  & 256           & RNN           & 24             & 0.6 \\ \hline
% \end{tabular}
% \vspace{-2.5ex}
% \end{table}

\section{Data}

We used the MAHNOB Mimicry Database \cite{bilakhia2015mahnob} to train and evaluate the systems in this paper. It contains 11.5 hours of spontaneous dyadic conversations on different topics. The purpose of the corpus was to be able to study dyadic mimicry behavior. The data-gathering used a setup of 15 shutter-level synchronized cameras, two close-talking microphones and one room-capturing microphone. The video streams capturing the faces were gray-scale. 40 participants discussed various subjects over 53 sessions (originally 54 sessions, but one session did not contain data for both participants). The average session length was $13\pm3.5$ minutes. 40 sessions of this dataset have additionally been annotated with mimicry episodes and occasionally their strength. For selecting mimicry segments for the evaluation we used segments annotated for smile, head nod and laughter. For more information, please refer to the original publication \cite{bilakhia2015mahnob}. 
The data was partitioned into an even split of one minute long, randomly-selected segments. We split the dataset in the following way: train 83\%, val 10\% and test 6.5\%. Additionally, one full session was held out completely (the remaining 0.5\%).

% frown, head nod, 'head_pitch', 'head_roll', 'head_shake', 'head_throwback', 'head_waggle', 'head_yaw', 'laughter', 'smile', 'surprise']

\subsection{Feature extraction}
\label{sec:feature_extraction}
From the videos (one camera angle per person and session) we extracted 2,068,410 image frames at 25 fps. OpenFace \cite{baltrusaitis2018openface} was then used in order to extract facial landmarks, which were used to determine bounding boxes for cropping and for the FLAME fitting. Cropped images were fed into RingNet \cite{RingNet:CVPR:2019} to estimate initial FLAME parameters. 
The RingNet output together with the facial landmarks were passed into the FLAME fitter in order to determine the final FLAME parameters, which were obtained through two optimizations outlined in \cite{FLAME:SiggraphAsia2017}. The result was a 100D PCA expression vector, a 12D pose vector with rotations, and a 300D PCA shape vector. From the expression vector we used the 50 first components together with the neck (3D) and jaw (3D) rotations to form our \textbf{facial features} (56D). Lastly some temporal smoothing was applied using Savitzky-Golay filtering (window length = 9, polynomial order = 3).

% fittings using the L-BFGS-B optimization algorithm. The first optimization tried to roughly fit the facial landmarks obtained from the image to facial landmarks associated with vertices in a template face model in the FLAME topology by minimizing scale, translation and rotation. The second optimization further tried to fit the landmarks, but now including regularizers in the cost function and the flame parameters controlling the template facial model.
From the audio we extracted 25 MFCCs + 1 log total frame energy (window length = 0.02 s, step size = 0.01 s, nfft = 1024) using python-speech-features \cite{pythonspeechfeatures}.
%\footnote{\href{https://python-speech-features.readthedocs.io}{https://python-speech-features.readthedocs.io}}
Additionally we extracted prosodic features (pitch, pitch delta, energy, and energy delta) using Praat \cite{boersma2011praat}.
%\footnote{https://www.praat.org}
The MFCCs and prosodic features were concatenated in order to create the \textbf{acoustic features} (30D). 

As some of these processing steps were computationally demanding (measurable in CPU+GPU months), the  extracted features are publicly available from the project website.

\begin{figure} %[hb]
\begin{center}
\includegraphics[width=0.4\textwidth]{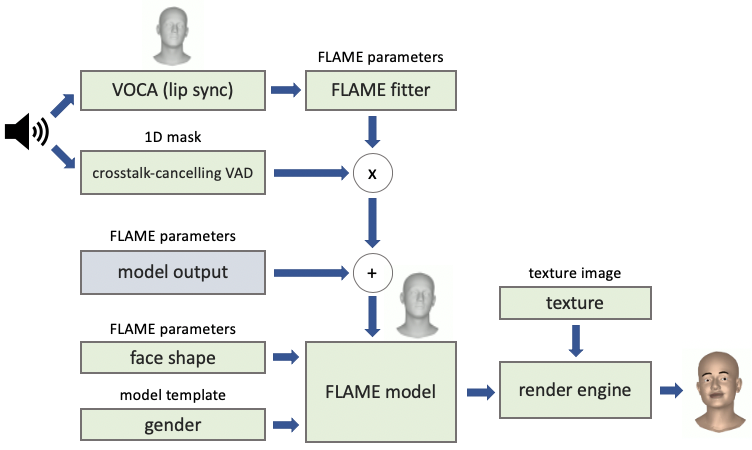}
\vspace{-0.75ex}
\caption{Stimulus generation pipeline, showing how audio is transformed into lip motion and then combined with the model output and rendered.}
% \vspace{-2ex}
\label{fig:stimuli_generation}
\end{center}
% \vspace{-2mm}
\end{figure}

\subsection{Stimulus-generation pipeline}
\label{sec:stimuli_generation}
A number of processing steps, illustrated in Figure~\ref{fig:stimuli_generation}, were necessary to generate the video stimuli:
First, Voca \cite{VOCA2019} was used to generate lipsync for all audio within the test segments. Voca takes audio as an input and outputs vertices in the FLAME topology. A template mesh was then fitted to these vertices using the method described in \cite{FLAME:SiggraphAsia2017} in order to obtain FLAME-parameters for the expression and jaw parameters. 
A simple energy-based crosstalk \ac{VAD} was implemented to output a mask for canceling crosstalk between the two speakers. This mask was the same length as the number of frames of the FLAME-parameters for the lipsync and was multiplied with each lipsync track. The result was subsequently added together with the model output, resulting in an avatar whose lip-movements are driven by recorded agent speech but whose facial gestures can be generated and manipulated independently. A random gender and a random face shape were sampled in the face-shape parameter space and were, together with the previous output, passed to the FLAME model to obtain 3D vertices. The gender decides which template model the FLAME model will use, and can be generic, female or male. Finally the resulting vertices together with a random texture were passed to the rendering engine, here Pyrender\footnote{\href{https://github.com/mmatl/pyrender}{https://github.com/mmatl/pyrender}}.

\section{Evaluation}
\label{sec:evaluation}
In this section we describe the subjective  experiments we conducted, specifically an ablation study, and the complimentary objective measure used to evaluate our model.
We ablated several key components of the model, namely the modalities it used as input and the presence of the special training scheme with negative samples. The specific ablations we considered were: \textit{no-face}: model not conditioned on the interlocutor's facial features;
\textit{no-speech}: model not conditioned on the interlocutor's speech features;
\textit{no-neg-train}: model trained without the negative samples described in Section~\ref{sec:training_scheme}.
% \begin{itemize}
%     \item \textit{no-face}: model not conditioned on the interlocutors facial features
%     \item \textit{no-speech}: model not conditioned on the interlocutors speech features
%     \item \textit{no-neg-train}: model without negative samples in training (see Section~\ref{sec:training_scheme})
% \end{itemize} 
For each ablation we conducted a separate hyperparameter search on the validation dataset to find the optimal setup and re-trained the models from scratch using the best hyperparameters, to enable the most fair comparison. The exact hyperparameters for these models are provided on the project website.

The ablation study also evaluated how the models perform when they receive mismatched conditioning, to try to understand to what extent the models take the various multi-modal signals into account. We call the instances when the avatar's speech was taken from another context ``\textit{mismatched} $S^a$'', when the interlocutor's speech was from another context ``\textit{mismatched} $S^i$'', and when the interlocutor's facial gestures were from another context ``\textit{mismatched} $F^i$''.

\subsection{Subjective evaluation setup}
Five experiments were carried out on Amazon Mechanical Turk (AMT) to evaluate human perception of the produced facial gestures.
%The studies were all carried out on .
The five experiments were designed to answer the following five questions: (1) Can participants discern appropriate facial gestures using our visualization? % GT vs deranged -
(2) Does our model take interlocutor input into consideration? % Prop vs Prop random align; 
(3) What is the importance of the interlocutor's facial features as input?
(4) What is the importance of the interlocutor's speech features as input? 
% (3) What multi-modal input was most important, speech or face (evaluating face)? % prop vs prop - p2 face; 
% (4) What multi-modal input was most important, speech or face (evaluating speech)? %Prop vs prop - p2 speech; What multimodal input was most important, speech or face?
(5) Does the training scheme with negative samples significantly improve the perceptual quality of output gestures?

%In order to tease out the contribution of each modality from the interlocutor we ran two experiments were each of them had been removed.

%Our analysis was consistently done in a double-blind fashion such that the conditions were obfuscated during analysis and only revealed to the authors after the statistical tests had been performed.

\subsubsection{Procedure}
The procedure of our experiment was similar to that described in \cite{feng2017learn2smile}. Every participant was first provided instructions and then completed a training phase to familiarize themselves with the task and interface. The training consisted of three items showing the participants what kind of videos they may encounter during the study. Each participant was then asked to evaluate video pairs. In all studies participants compared two videos, each containing two virtual characters interacting with each other (see Figure~\ref{fig:intro}). The participants were always asked to only evaluate the avatar on the right, since it was the only one that was manipulated; the left avatar -- the interlocutor -- was always the same between both videos, and its movements reflected the same segment of ground-truth motion in the data. The videos were presented side by side and could be replayed separately as many times as desired. For each pair, participants indicated which video they thought best corresponded to the given question and there was also an option to state that they perceived both videos to be equally appropriate. The question we asked was always the same across experiments and similar to that used by Ahuja et al.~\cite{ahuja2019react}:
``\textit{Which of the two characters on the right side of each video has the most appropriate behavior in response to the character on its left?}''

All subjective tests used a binomial sign test with Bonferroni correction for the five studies. Ties were excluded.

%, and was chosen to be the interlocutor which speaks more using the sum of voice activity from the VAD.

\subsubsection{Stimuli}
Since the goal was to evaluate facial gestures, audio was removed, but lip-sync, based on the original audio for each character, was retained and was the same between both videos in each evaluated pair.
This choice was based on other facial gesture studies such as \cite{feng2017learn2smile} and on the fact that an informal pre-study (12 participants), found that participants tended to base their judgments on how well the motions matched the semantic content, rather than the interlocutor interaction.
We found this to be inappropriate for our study since no explicit linguistic understanding was built into our model.
%found that subjects exposed to speech information tended to ascribe semantic significance to the generated gestures, 
The avatars were placed side by side and facing forward, adjusted such that the 3D avatar would face the viewer when the original talker was facing the other interlocutor in the original interaction. Neck rotation was subtracted from the eyes, giving the impression of the avatar looking straight at the viewer even when turning its head. Head shape, gender and skin color (see Figure~\ref{fig:intro} for an example) were randomized but kept constant for each video segment across all experiments. Which conversation party from the original ground-truth interaction that was selected as the interlocutor and placed on the left was based on who spoke the most in that segment, determined by summing the \ac{VAD} output.

22 video pairs were evaluated in each experiment, except for Experiment 1, where 64 video pairs were evaluated (34 mimicry and 30 non-mimicry segments) and each participant evaluated 10 random pairs of each type. Segments were randomly counterbalanced and (like the original mimicry annotations) varied in duration ($3\pm{}2$~s) from one to eight seconds. All experiments used the same segments, except Experiment 1 which had additional segments as above.

A few randomly-selected examples generated by our method and used for the experiments are available on the project website. Since some of these sequences were jittery, we also provide examples where we ``lowered the temperature'' of the underlying Gaussian (we set $\sigma^2<1$ for \textbf{Z}) \cite{kingma2018glow}, which produced smoother motion.
We did not apply any smoothing filters to the output in this work.

\subsubsection{Participants}
All participants were recruited through AMT and were only allowed to participate once in any of the studies. The participants had to have an acceptance rate of at least 98\% and completed over 10,000 previous HITs to be eligible for our study.
We used attention checks to filter out inattentive participants.
For two of the attention checks (one early in the experiment, one close to the end) we added a text telling the participant to report the video as broken.
%One of them was positioned early in the experiment and another close to the end.
Participants were excluded if they failed any of these attention checks.
The other three attention checks comprised pairs presenting the exact same video twice and were placed at the 7th, 10th, and the 15th trial-position for all experimental sessions. Here, an attentive rater should answer ``no difference''. Participants were excluded if they failed all three of these attention checks. 

\subsection{Results of subjective evaluation}
\begin{figure}[t] %[ht]
\begin{center}
\includegraphics[width=0.33\textwidth]{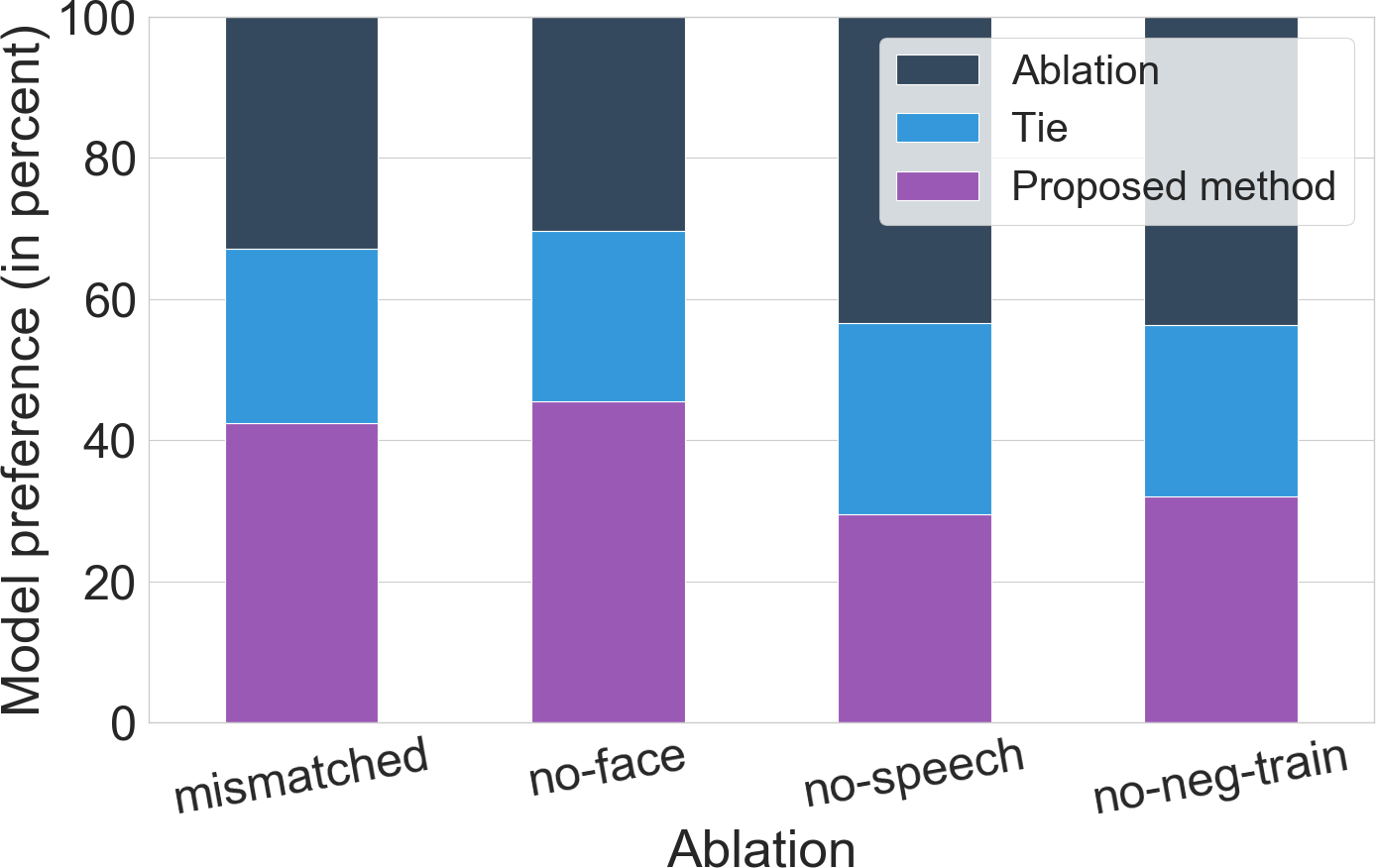}
% \vspace{-0.75ex}
\caption{Results from the subjective ablation studies.}
% \vspace{-2ex}
\label{fig:result2r}
\end{center}
% \vspace{-2ex}
\end{figure}

The results for Experiment 2 (\textit{mismatched}), 3 (\textit{no-face}), 4 (\textit{no-speech}), and 5 (\textit{no-neg-train}) are shown in Figure~\ref{fig:result2r}.

\subsubsection{Experiment 1: Matched and mismatched ground truth}
First we evaluated if our stimulus-generation methods allowed online workers to perceive a difference between the actual facial gestures (\textit{ground truth} condition) and avatar gestures taken from another point in time in the same interaction but with the same person (\textit{mismatched} condition).
We recruited 30 participants (14 female, 16 male), all from the USA. Their mean age was 37.4 with a std of 11.1. %Seven participants were excluded from the study due failing attention checks (four) or stopping the study (three).

We conducted a binomial sign test with Bonferroni correction excluding ties to analyze the responses separately for the two types of stimuli: the mimicry segments and the non-mimicry segments.
The ground truth videos were preferred over the Mismatched ones for mimicry segments ($p<0.001$). There was no statistical significance for the non-mimicry segments ($p$=1). These results indicate that online workers can indeed distinguish the Mismatched facial gestures from the ground truth, but only in segments where that difference is salient, e.g., if the conversation parties display strong non-verbal interactions such as mimicry.
%the difference is evident, such as when mimicry is happening. 
Given this result we concentrated our remaining evaluations on mimicry segments, since they provided for the clearest distinction between appropriate and inappropriate agent behavior.
As the non-mimicry segments did not produce a statistical difference they were excluded from
%this and the following
remaining studies.
Furthermore, since our model required 24 frames (0.96 s) of initialization data, only 22 samples could be used for the remaining experiments.

%Our first experiment on ground-truth data identified segments where the interlocutor invites to mimicry behavior as locations where it is particularly important for a non-verbal agent to be responsive to the interlocutor. We therefore concentrated our remaining evaluations on these segments, since they provided for the clearest distinction between appropriate and inappropriate agent behavior."

\subsubsection{Experiment 2: Matched and mismatched proposed model}
\label{sec:prop_vs_rand}
In the second experiment we evaluated whether the proposed model actually uses the interlocutor's input when generating facial gestures. To this end, we shuffled the conditioning information like before, creating mismatched stimuli where the conditioning information from the interlocutor was always taken from a different sample than the motion used by the interlocutor avatar in the video (but still from the same session and the same person). We compared the proposed model's facial gestures using normal test sequences versus those using mismatched sequences.
%If the model is able to take the interlocutor's behavior into account to generate more appropriate behavior, this could in a difference .
This use of matched and mismatched samples has the advantage that the quality of the motion is the same across the conditions seen in the videos (since all avatar motion was generated from the same trained model); only the appropriateness of the motion may differ between the two.

We recruited 30 participants (22 male, 8 female). The majority (29) were from the USA. Their mean age was 33.7 with a std of 6.9. The test showed a statistically significant difference between the model output on matched and mismatched test sequences. Specifically, there was a preference towards the matched sequences ($p=0.032$).

\subsubsection{Experiment 3: Ablating facial gestures}
\label{sec:no_face_experimentt}
% - Proposed model vs proposed model w/o $F^i$
Here we compared the proposed model (\textit{proposed} condition) against the ablation where the interlocutor's facial gestures was not available to the model (\textit{no-face} condition). We recruited 30 participants (19 male, 10 female, 1 non-binary), of which 29 were from the USA. The mean age was 37.3 with a std of 9.4. The test showed a statistically significant preference for the proposed model over the \textit{no-face} ablation ($p<0.001$).

\subsubsection{Experiment 4: Ablating speech}
\label{sec:no_speech_experiment}
%Proposed model vs proposed model w/o $S^i$
In this experiment we compared the proposed model (\textit{proposed} condition) against the ablation where the interlocutor's speech was not available to the model (\textit{no-speech} condition). We recruited 30 participants (16 male, 13 female, 1 non-binary), of which 29 were from the USA. Their mean age was 36.6 with a std of 8.7. The test showed a statistically significant preference for the \textit{no-speech} ablation ($p<0.001$).

\subsubsection{Experiment 5: Negative sample training}

In this experiment we compared the proposed model (\textit{proposed} condition) against the same model without any negative samples during training (\textit{no-negative-training} condition).
%no special training scheme was applied

We recruited 30 participants (17 male, 13 female), of which 28 were from the USA. Their mean age was 38.7 with a std of 12.6. The test showed a statistically significant preference for the model trained without the special training scheme ($p=0.003$).

\subsection{Objective evaluation}
It is difficult to evaluate the quality of facial gestures objectively, and it is even harder to objectively evaluate whether or not facial gestures are adapted to the interlocutor. Calculating distance from recorded ``ground truth'' motion is not meaningful, as a multitude of different gestures can be appropriate even if the conditioning input is fixed. We instead considered the likelihood since normalizing flows enable direct probabilistic inference, letting us calculate the log-likelihood of test data under our model. The test data should have high likelihood only if we model the data distribution well. We evaluated log-likelihood for the proposed model and its ablations for unmodified test sequences as well as mismatched sequences as defined above. The average values along with their standard deviations are given in Table \ref{tab:nll_results}. The interpretation of the results is discussed in Section~\ref{sec:discussion}.
\begin{table}[!t]
\footnotesize
\caption{Log-Likelihoods for the proposed model and its ablations on test sequences and mismatched versions thereof.}
\vspace{-0.75ex}
\label{tab:nll_results}
\begin{tabular}{@{}lllll@{}}
\hline
 System                 & All correct       & mismatched $\boldsymbol{S^a}$   & mismatched $\boldsymbol{S^i}$   & mismatched $\boldsymbol{F^i}$   \\
\hline
 Proposed             & 40051$\pm$144       & 40050$\pm$144                   & 40050$\pm$\hphantom{000}144                   & 23522$\pm$99436                \\
 no-face                & 38141$\pm$240       & 38141$\pm$238                   & 31614$\pm$144323                & -                           \\
 no-speech              & 35545$\pm$\hphantom{0}67        & 35544$\pm$\hphantom{0}68                    & -                             & 35538$\pm$\hphantom{000}68                   \\
 no-neg-train           & 38698$\pm$\hphantom{0}92        & 38698$\pm$\hphantom{0}93                    & 38699$\pm$\hphantom{0000}92                    & 38654$\pm$\hphantom{000}97                   \\

\end{tabular}
\vspace{-3ex}
\end{table}

\section{Discussion}
\label{sec:discussion}
The purpose of Experiment 2 (Section~\ref{sec:prop_vs_rand}) was to see if our method can leverage the multi-modal input to generate more appropriate motion in response to the interlocutor. We found a significant preference for when the model outputs facial gestures relevant to the context, as opposed to a random context, indicating that we successfully generated interlocutor-aware facial gestures. This result is in line with the findings from Experiment 1, where it was shown that evaluators can indeed distinguish -- and furthermore prefer -- non-verbal behavior which is dependent on the interlocutor over any random (coherent) facial gestures.

Experiments 3 and 4 were designed to assess the relative importance of different interlocutor input modalities.
Experiment 3 (Section~\ref{sec:no_face_experimentt}) considered removing the interlocutor facial information.
This made the model perceptually significantly worse.
In addition, this \textit{no-face} condition
%gave the jerkiest motion (Table~\ref{tab:jerk_results}) and
gave likelihoods that were significantly affected by mismatched speech information (Table~\ref{tab:nll_results}), suggesting that, lacking facial information, the model instead became more attuned to the interlocutor's speech, possibly to the point of overfitting.

%Furthermore, removing the facial information from the interlocutor, as investigated in Experiment 3 (Section~\ref{sec:no_face_experimentt}), yielded a model which is perceptually significantly worse than when the model is provided the facial gestures. Also worth noting is that the likelihood for when the model is provided mismatched interlocutor speech seem to be heavily affected, suggesting that the model, lacking information about the interlocutors facial information, instead becomes more attuned to the speech information. This model also rendered the jerkiest movements, as can be seen in Table~\ref{tab:jerk_results}.

If we instead removed the interlocutor speech input (Experiment 4, in Section~\ref{sec:no_speech_experiment}), the resulting ablation performed significantly better than the proposed model.
%However, removing the speech, which Experiment 4 (Section~\ref{sec:no_speech_experiment}) investigated shows that the model performed significantly better than the proposed model.
This suggests that the facial information is the most important for the model, at least in this no-audio evaluation paradigm. It is surprising that the model with facial information alone was better than the one using face and speech together. Speculatively, this might be due to the type of speech features used, and experimenting with less speaker-dependent speech representations would be interesting for future work. %Additionally, we can see that both the \textit{no-speech} and \textit{no-neg-train} conditions had lower range of motion than the ground truth, indicating that they were not as animated and closer to the mean face. While being closer to the mean face might yield a less expressive and animated result, it also provides less room for making mistakes that the evaluators are sensitive to.
%Finally, it is important to note that there was no audio in the evaluated stimuli, which may have affected the results.

There is an intriguing disparity between the likelihood numbers in Table~\ref{tab:nll_results} -- where negative training helped models learn to more effectively assign probability mass to motions matching the interlocutor (as opposed to non-matching motion) -- and the subjective results from Experiment 5, which found that not using negative samples in the training was perceived significantly better.
While negatively-trained models clearly were able to learn to distinguish well between scenarios with matched and mismatched modalities, they do not appear to have leveraged this to generate more appropriate motion in matched setups.
However, it is also well known that likelihoods and human ratings are sensitive to different modeling aspects (see, for instance, \cite{theis2016note}). Thus higher likelihood does not necessarily mean better perceptual quality, and our findings here are likely another reflection of that fact.

A potential limitation of this work is the fact that we are evaluating multi-modal interactions that contain speech, but without revealing that speech to the evaluators. This was a deliberate choice, as we in a pre-study on mismatched ground-truth motion found that participants otherwise tend to assign an inordinate significance to the linguistic content and how the avatar moves and behaves in relation to that content. Since the presented method does not attempt to model semantics, removing the speech would make it less likely that evaluators assign spurious semantic meaning to the gestures, and instead force them to evaluate the motion in a non-semantic way.
It is also consistent with previous evaluation of non-verbal facial gestures, e.g., \cite{feng2017learn2smile}.
%We furthermore replicated Experiment 2 presented in Section~\ref{sec:prop_vs_rand} with the presence of speech stimuli (n=30), finding a statistical difference (p<0.05) that agrees with the result presented above, although the effect is smaller, providing evidence that the evaluation is more difficult when speech with semantic content is present.
Furthermore, we replicated Experiment 2 from Section~\ref{sec:prop_vs_rand} with n=30 subjects, but with speech audio present in the video stimuli. We found a statistical difference (p<0.05) in agreement with Experiment 2, but the effect was less significant (0.04998), supporting the pre-study finding that the presence of speech with semantic content confounds the evaluation of the non-verbal facial gestures.
%
%would focus the evaluators to not attach any semantic meaning to the gestures, and instead evaluating them in a non-semantic way.
%Additionally, in Experiment 1 where we evaluated ground truth compared to mismatched ground truth segments, evaluators would have been able to guess which segment based on the audio, if there was for example laughter or verbal references to any actions in any of the videos.
In general, we believe that the absence of speech audio would be most likely to affect evaluators' assessments of the impact of the speech modalities on the motion, such as the results of Experiment 4. Another limitation is that the evaluated segments, being annotated mimicry episodes, were rather short. In some cases, they may then be considered hard to evaluate.
%In some cases they were considered hard to evaluate.
%The segments were chosen based on the annotated mimicry segments and to be non-overlapping.
%Another limitation of our work is that we only used mimicry segments in our subjective ablation study. The reason for choosing only these segments was that participants were able to distinguish between the ground truth and mismatched samples  It would be valuable to evaluate adaptive facial gestures in broader scenarios in the future.
\vspace{-3.5ex}
\section{Conclusion}
We have presented a method for probabilistic and interlocutor-aware facial-gesture generation based on multi-modal inputs.
Experiments found that human raters significantly preferred facial gestures generated in response to the interlocutor over mismatched facial gestures that did not take the interlocutor into account.
This shows that the proposed approach managed to leverage the multi-modal input to generate better gestures. We evaluated our system on mimicry segments due to their perceptual saliency, but it should be stressed that no information relating specifically to mimicry was used during training.
%Our experiments found a significant preference among human raters for facial gestures generated given the correct interlocutor context for the given interaction showing that the model does leverage the multi-modal input. 
The subjective appropriateness of generated motion decreased significantly when information about the interlocutor's facial gestures was omitted, suggesting that this modality is of major importance to the task.

Future work should investigate the use of other parametrizations of multi-modal signals, especially speech representations, and various ways of incorporating them into the model.
%Especially the speech representations should be investigated further.
It would also be highly interesting to investigate how this method would work in a real-time interaction with a user.

\begin{acks}
The authors acknowledge the support from the Swedish Foundation for Strategic Research, project EACare under Grant No.: RIT15-0107 and the Wallenberg AI, Autonomous Systems and Software Program (WASP) funded by the Knut and Alice Wallenberg Foundation.
(Portions of) the research in this paper uses the MAHNOB MHI Mimicry database collected by Prof.\ Pantic and her team at Imperial College London, and in part collected in collaboration with Prof.\ Nijholt and his team of University of Twente, in the scope of MAHNOB  project financially supported by the European Research Council under the European Community's 7th Framework  Programme  (FP7/2007-2013) / ERC Starting Grant agreement No.\ 203143.
% \cite{jonell2017machine}
\end{acks}

\bibliographystyle{ACM-Reference-Format}
\bibliography{refs}

\end{document}